# Short-term Mortality Prediction for Elderly Patients Using Medicare Claims Data

Maggie Makar, Marzyeh Ghassemi, David M. Cutler, and Ziad Obermeyer

*Abstract*—Risk prediction is central to both clinical medicine and public health. While many machine learning models have been developed to predict mortality, they are rarely applied in the clinical literature, where classification tasks typically rely on logistic regression. One reason for this is that existing machine learning models often seek to optimize predictions by incorporating features that are not present in the databases readily available to providers and policy makers, limiting generalizability and implementation. Here we tested a number of machine learning classifiers for prediction of six-month mortality in a population of elderly Medicare beneficiaries, using an administrative claims database of the kind available to the majority of health care payers and providers. We show that machine learning classifiers substantially outperform current widely-used methods of risk prediction—but only when used with an improved feature set incorporating insights from clinical medicine, developed for this study. Our work has applications to supporting patient and provider decision making at the end of life, as well as population health-oriented efforts to identify patients at high risk of poor outcomes.

## I. INTRODUCTION

Physicians and patients routinely face important decisions regarding medical care at the end of life, ranging from aggressive curative treatment to comfort-oriented palliative care. Such treatment decisions depend heavily on a patient's expected survival, or prognosis: high-risk interventions for example surgery or chemotherapy may be reasonable options for patients with longer expected survival, but are less well-suited to patients with poor prognoses—who assume all the risks of the intervention but are unlikely to live long enough to realize its potential benefits. The adoption of the Affordable Care Act has heightened interest in strategies to target poor-prognosis patients for interventions to improve quality of care at the end of life, and at the same time reduce costs. Thus the availability of accurate prediction tools is important to support both physician and patient decision making at the end of life, as well as policy-level population health management efforts.

Despite the importance of accurate risk prediction, there are few examples of mortality prediction models being used in a real-life clinical or policy setting. This is for at least two reasons. First, current approaches to identification of high-risk patients in the clinical literature and classification tasks in general rely on logistic regression to predict mortality, mostly using small datasets and small feature sets [1]. Besides producing biased coefficients, this on average underestimates predicted probability of rare outcomes like death [2]. In addition, several basic assumptions of logistic regression are problematic in the high dimensional, highly correlated feature sets needed to accurately characterize medical illnesses, including illnesses, health care utilization, and demographics.

Second, while the machine learning (ML) literature contains many tools for predicting clinical outcomes, these typically rely heavily on feature sets derived from test results or *de novo* data collection from patients—examples include genomic data [3], vital signs [4], laboratory tests [5] or medical chart text [6]—which are typically not available in databases used by clinicians and policy makers, resulting in a technically excellent classifier with no clear path to implementation in a real health care environment. Conversely, despite the obvious practical advantages of using routinely collected data available to most health care providers with respect to implementation, administrative data have a poor track record of success for risk prediction. Indeed, models using these administrative data generate such poor predictions that a recent review in a leading medical journal could not recommend them for predicting outcomes in populations of older adults [1].

We set out to explore the performance of several ML classifiers for the prediction of short-term (six month) survival with administrative health data, building on insights from prior studies utilizing ML techniques with administrative data to answer similar questions [7], [8]. We hypothesized that some of the limitations of prediction using administrative data could be overcome with careful pre-processing and engineering of features, drawing on insights from machine learning [9] and clinical medicine [10] literature. We thus constructed an augmented set of variables capturing patients' comorbidity (disease burden), healthcare utilization, and functional status (ability to live and function independently). We tested these augmented features with a range of commonly used ML algorithms, in combination with class balancing techniques, and evaluated their performance compared to traditional measures of comorbidity commonly used to predict mortality in similar datasets. Our work could be used to aid physician and patient decisions regarding end of life care. On an aggregate population level, it could also serve to identify patients at a high risk of poor outcomes, as an input to efforts to manage and improve the health of populations.

Manuscript received October 15, 2014. This work was supported by NIH Common Fund grant, DP5 OD012161 (PI: Obermeyer).

M. Makar is with the Department of Emergency Medicine at Brigham & Women's Hospital, Boston, MA (e-mail: mmakar@partners.org).

M. Ghassemi is with the Department of Electrical Engineering and Computer Science, Massachusetts Institute of Technology, Cambridge (e-mail: mghassem@gmail.com)

D. Cutler is with the Department of Economics at Harvard University, Cambridge, MA, and National Bureau of Economic Research, Cambridge, MA (e-mail: dcutler@fas.harvard.edu).

Z. Obermeyer is with Department of Emergency Medicine at Harvard Medical School and the Department of Emergency Medicine at Brigham and Women's Hospital, Boston, MA (e-mail: zobermeyer@partners.org).

## II. DATA

We used Medicare claims data, which capture patient demographics, healthcare utilization, and recorded diagnoses from the time of Medicare enrollment till death, to develop our model. We used a nationally-representative 5% sample of all Medicare fee-for-service beneficiaries in 2010 and retained those over 65 years old living in the continental United States. We included only those alive and not enrolled in the Medicare hospice benefit (implying known terminal disease) on an arbitrary date, $t_0$ (July 1, 2010), to mimic a real life situation in which predictions are needed at an arbitrary point in time.

We segmented the population into non-mutually exclusive cohorts based on presence of major individual diseases, as traditionally defined in the health services research literature [10], based on whether or not a patient was diagnosed with the disease in the year prior to $t_0$. We selected four disease cohorts—congestive heart failure (CHF), dementia, chronic obstructive pulmonary disease (COPD), and any tumor, all major causes of death and disability nationally—to explore how predictions vary across a range of underlying medical pathologies and mortality rates. We randomly chose 20,000 beneficiaries in each disease group from the national sample (of approximately 2, 0.8, 2.7, 2.3 million, respectively), and split each equally into development and validation datasets.

We used death dates from the Medicare data to define our outcome of interest, death in the six months after $t_0$. Six-month survival is the major eligibility criterion for the hospice benefit, making it a useful interval for both physicians and patients to start considering the goals of end of life care. We used a one-year look-back period before $t_0$ to create features for predicting death. Over this period, we extracted information about patient demographics, healthcare utilization, comorbidity, durable medical equipment and medical diagnoses using inpatient and outpatient, home health and durable medical equipment claims.

### A. Traditional Model and Variables

In order to compare the performance of our final classifier to existing practices, we used a model developed by Gagne et al [10] which was identified by a recent review as the best prognostic tool [1] for predicting mortality using administrative data. This model used age, sex, and a set of 20 indicator variables corresponding to individual diseases or conditions (a combination of the commonly-used Elixhauser and Charlson indices).[1] Indicators were initially set to zero. If an International Classification of Disease (ICD) code corresponding to any of these comorbidities was present on any claim in the one-year look-back period before $t_0$ the indicator was set to one, accounting for outpatient 'rule-out' diagnoses as is usual [11].

### B. Augmented Feature Set Creation

We set out to create an 'augmented' set of clinically relevant health measures that captured disease severity and progression with time as well as functional status.

In order to capture disease progression, we replaced each of the 20 comorbidity indicator variables from the traditional model with two count variables representing the total number of medical encounters (such as clinic or emergency room visits and inpatient or skilled nursing facility stays) involving that comorbidity, in two time periods: 1-3 months, and 3-12 months prior to $t_0$. Unlike traditional indicator variables these features capture not only presence or absence of a disease, but its severity (measured by number of encounters) and evolution over time (captured by two features for two discrete time periods, a recent period and a baseline period).

Figure 1 shows the difference between the traditional and the augmented variable sets, for two hypothetical patients with different trajectories of disease $x$ over the look-back period. Ticks on the $x$-axis show the individual patient–physician encounters related to disease $x$, dashed and solid lines show the values of the traditional and augmented variables respectively. In this scenario, Patient 1 experienced progressively worsening disease while Patient 2's disease was treated or resolved. These two patients would be have the same value on a traditional measure of disease $x$, despite clear difference in their disease trajectory—and likely prognosis—while the augmented variables capture the two different trajectories. We initially chose these specific time periods (1-3 and 3-12) based on visually inspecting healthcare utilization patterns, and observing a sharp increase in the number of total diagnoses three months before death (see Figures 2 & 3). We later empirically validated our choice as outlined in the analysis section.

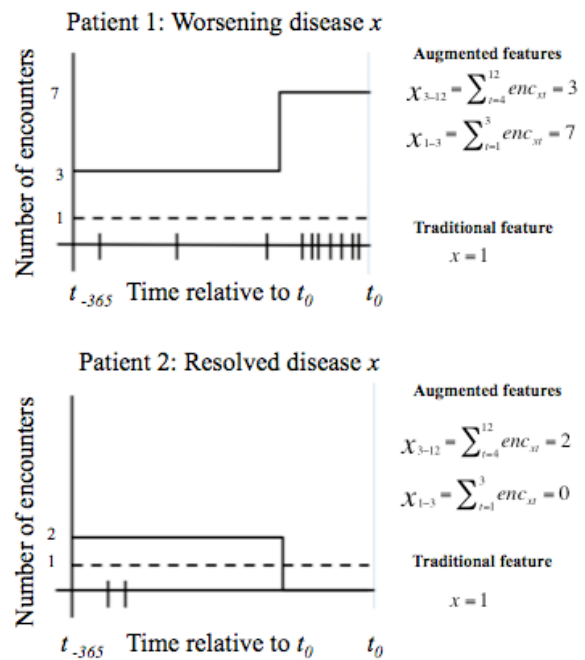

Fig. 1. Augmented vs. traditional variables

We also developed a new set of features measuring functional status, a patient's ability to live and function independently and without assistance. Functional status is known to have considerable prognostic importance [12], [13] but does not figure into any existing set of comorbidities derived from administrative data. We identified claims for durable medical equipment (e.g., walker, wheelchair, home oxygen) as well as non-specific ICD codes (e.g., 728.87 muscle weakness, general; 783.7 failure to thrive-adult). Each of these measures was included as count variables, split into 1-3 and 3-12 time periods. We also included counts of utilization of health care services including clinic and ED visits, inpatient hospitalizations, home health assistance, and skilled nursing facility for each of the two time periods. Finally, we added income and race, which have previously been identified as predictors of mortality [14], likely because they account for unmeasured differences in baseline illness and access to care. The full list of variables is included in Table 1.

---
[1] The original paper by Gagne et al. [10] used a sample from a single US state, whereas we use a sample drawn from across the US. We included a vector of 51 state indicator variables for each state in continental US as well as D.C. to account for this difference.

TABLE I. COMPARISON BETWEEN VARIABLES IN THE AUGMENTED AND TRADITIONAL VARIABLE SETS †

|  | Traditional | Augmented |
|---|---|---|
| **Patient demographics** | | |
| Age | + | + |
| Sex | + | + |
| State of residence | + | + |
| Race | - | + |
| Income deciles | - | + |
| **Disease groups:** Alcohol abuse, deficiency anemia, cardiac arrhythmias, coagulopathy, complicated diabetes, dementia, fluid & electrolyte disorders, hemiplegia, HIV/AIDS, hypertension, liver disease, metastatic cancer, CHF, psychosis, COPD, chronic pulmonary disease, peripheral vascular disorder, renal failure, tumor and weight loss | Dummy variable | Two count variables for each disease group: one for 1-3 and another for 3-12 months before July 1st |
| **Functional status:** Aerodigestive-GI/GU, aerodigestive-nutrition, aerodigestive-respiratory, general weakness, general mobility, cognition, sensorium and speech | - | Two count variables for each group: one for 1-3 and another for 3-12 months before July 1st |
| **Durable medical equipment:** bed, cane/walker, $O_2$ and wheelchair | - | Two count variables for each durable medical equipment: one for 1-3 and another for 3-12 months before July 1st |
| **Utilization\*:** Home health days, inpatient days, skilled nursing facility admissions, inpatient admissions, clinic and emergency room visits | - | Two count variables for each encounter: one for 1-3 and another for 3-12 months before July 1st |

† The plus sign (+) signifies presence while the minus sign (-) signifies absence
\*Skilled nursing facility stays were represented by two indicator variables, one for 1-3 and another for 3-12 months before July 1st, instead of count variables due to limitations of claims information on these encounters
CHF denotes congestive heart failure
COPD denotes chronic obstructive pulmonary disease
ICD denotes International Classification of Disease codes
GI/GU denotes gastrointestinal and genitourinary impairments

III. ANALYSIS

Since few papers have applied ML to administrative claims data, we first surveyed current literature to identify the most common supervised learning techniques used in similar settings with high-dimensional, highly correlated features, and an imbalanced binary outcome [15], [16]. We implemented six classifiers: naïve Bayes, support vector machines (SVM), K nearest neighbors (k-NN), artificial neural nets (ANN), random forests (RF) and logistic regression, both with L1 regularization (lasso) and without.[2] We analyzed each of the four disease cohorts separately, using the same development and validation samples to compare the different classifiers.

We chose area under receiver operating curve[3] (AUC) as our primary performance measure for all classifiers. Given that six-month mortality was a rare outcome in all four disease cohorts (10% or under), optimizing accuracy could have resulted in a fairly accurate model that simply predicted that all subjects would live. In all cases, we split the development dataset into two randomly chosen mutually exclusive samples of equal size. The first was used to train models ('training sample') while the second was used to tune the parameters ('tuning sample'). We report results only for the validation dataset.

For each of the six classifiers, we identified the optimal combination of tuning parameters using a greedy search algorithm. We first trained the model using the training sample setting all the model parameters to their default values, then changed one parameter at a time and chose the value that maximized AUC of the tuning sample. We used the optimal value when tuning the second parameter and so forth. Second, given class imbalance in this dataset, we experimented with several up- and down-sampling strategies in order to present the classifier with a balanced dataset of alive and dead patients: up-sampling, creating duplicates of the minority class, and down-sampling, choosing a random sample of the non-events, both resulting in class balance in the final training dataset. In each of these cases, we up-sampled or down-sampled the training dataset and used the original tuning dataset to test the performance of the classifier. We repeated this process for each of the four disease cohorts and for each variable set; the traditional and the augmented.

For each method and each disease cohort we created the model using the development dataset then predicted the outcomes for the 10,000 beneficiaries in the validation dataset and finally calculated the AUC as well as a bootstrapped standard deviation using 1,000 replicates. We identified the best performing model by identifying the maximum AUC for each method and each disease cohort. After identifying the best performing classifier, we empirically validated our choice to discretize the augmented variables into 1-3 and 3-12 months before $t_0$. Using the same development and validation samples, we created 11 alternate variable sets, similar in principle to the augmented variables described above, but discretizing over different time periods: 1-2 months and 2-12 months, 1-4 and 4-12 and so forth (one of these sets was simply a vector of count variables summed over months 1-12). We then trained the best classifier using the same training samples with these 11 additional variable sets and used predictions from the validation sample to calculate the AUC.

IV. RESULTS

Mortality rates in the four development samples were 4.7%, 4.9%, 7.7% and 10.1%, for the tumor, COPD, CHF and dementia cohorts respectively. Rates in the validation samples were similar: 4.4%, 4.8%, 7.7%, and 9.7%.

---

[2] All analyses were done using R version 3.1.0. Naïve bayes classifiers were built using the klaR package, SVMs using the kernlab package, k-NNs using the caret package, ANN using the nnets package, RFs using the randomForest package, lasso using the glmnet package and finally logistic regressions using the stats package.
[3] The AUC reflects the probability that given two patients, one with positive outcome (death in the 6 months after July 1) and the other with a negative outcome (survival during the same period), the model will assign a higher probability of death to the former.

Table 2 shows the AUC (validation) using the augmented and traditional variables for each family of classifiers. We only present the results from the tumor group since relative performance of classifiers was largely similar across the four disease cohorts. Overall, the best performing model was the RF using the augmented variables, which had an AUC of 0.826 (SD = 0.010) using the pre-specified split on 0-3 and 3-12 months, and higher at an AUC of 0.828 (SD = 0.010) using the best split, which we determined retrospectively using validation data. All classifiers performed better when trained using the augmented variables. Logistic regression with L1 regularization was the best performing classifier using traditional variables.

The results in table 2 show the performance of the best model within each family of classifiers, from the validation dataset. For the random forest, the best model configuration used 1000 trees, with each split using sqrt($m$) variables (where $m$ is the total number of variables in the model, which varied between traditional and augmented sets) and a terminal node size of 1. We used a different randomly chosen balanced subsample of the data to grow each tree. The best lasso model was created using a down-sampled dataset for the augmented variables and the original sample using traditional variables. The original sample, without any re-sampling, gave the best results when training the naïve Bayes, regardless of which variable set was being used. The neural nets with 2 hidden layers, using a down-sampled dataset performed best with the traditional and augmented variables. We used 10 fold cross validation in the training dataset to estimate the best number of neighbors for the k-NN, which performed best with a down-sampled dataset for both variable sets. As for the SVM the best performing model used a radial basis function to map the data onto a higher dimension. The best sampling method for the logistic regression was down-sampling when using the augmented variables and up-sampling when using the traditional variables. Further data on the specific model fits is available from the authors on request.

TABLE II: VALIDATION AUC OF THE TUMOR COHORT ACROSS DIFFERENT CLASSIFIERS

| Classifier | Augmented variables | | Traditional variables | |
|---|---|---|---|---|
| | AUC | SD | AUC | SD |
| Random Forest | **0.826** | **0.010** | 0.774 | 0.012 |
| Lasso | 0.810 | 0.011 | 0.780 | 0.012 |
| Naïve bayes | 0.794 | 0.011 | 0.757 | 0.012 |
| Neural nets | 0.772 | 0.012 | 0.481 | 0.014 |
| kNN | 0.767 | 0.012 | 0.712 | 0.014 |
| SVM | 0.761 | 0.011 | 0.667 | 0.013 |
| Logistic regression | 0.563 | 0.053 | 0.727 | 0.014 |

Table II shows AUC and bootstrapped standard deviation of the tumor validation dataset for the best performing combination of tuning parameters and sampling techniques for each family of classifiers

Overall performance of all classifiers varied considerably across the four disease cohorts as shown in Table 3, which shows the validation AUC by disease cohort using the augmented variables split at 1-3 and 3-12 as well as the best split which varied by disease cohort. The tumor group had the highest AUC even though it suffered from the highest class imbalance. We tested the performance of the additional 11 variable sets that were created to test time period aggregations other than 1-3 and 3-12. We found that splitting the count variables into 1 month and 1-12 months produced the highest AUCs in the tumor (AUC= 0.828), COPD (AUC=0.814), and CHF (AUC=0.756) while splitting them into 1-3 and 3-12 periods (our initial split based on visual inspection) was best for the Dementia cohort (AUC=0.715).

TABLE III: RANDOM FOREST PERFORMANCE ACROSS DISEASE COHORT

| Cohort | Mort. rate (dev.) | Mort rate (valid.) | 1-3 and 3-12 split | | Best split* | |
|---|---|---|---|---|---|---|
| | | | AUC | SD | AUC | SD |
| Tumor | 4.7 | 4.4 | 0.826 | 0.010 | 0.828 | 0.010 |
| COPD | 4.9 | 4.8 | 0.811 | 0.009 | 0.814 | 0.009 |
| CHF | 7.7 | 7.7 | 0.752 | 0.009 | 0.756 | 0.009 |
| Dementia | 10.1 | 9.7 | 0.715 | 0.008 | 0.715 | 0.008 |

Table 3 shows AUC and bootstrapped standard deviation of the validation datasets using the random forest classified with augmented variables. "dev." denotes development (i.e., training and tuning subsamples), "valid." denotes validation, "Mort." denotes mortality.
* The best split for the tumor, COPD and CHF was 1 and 1-12 months while that of the Dementia was 1-3 and 3-12

These wide variations in AUC did not correlate with mortality rate—in fact, there appeared to be an inverse correlation, since the highest AUC was observed in the tumor cohort with the lowest mortality rate, and the lowest AUC in the dementia cohort with the highest mortality rate. This raised the possibility that certain aspects of the disease cohorts, unrelated to class imbalance, had large effects on model performance. We hypothesized that differences in underlying disease severity and evolution accounted for these differences. To measure this, we counted the number of unique diagnoses coded for each patient in the 12 months leading up to $t_0$, separated by outcome (dead or alive). Finally, we calculated the difference between the two means for each month. Figure 2 shows the mean number of coded diseases per patient on each month leading up to $t_0$. Patients who ultimately died are represented by solid lines while ones who did not are represented with dashed lines. Patients who ultimately died had a significant increase in the number of coded diseases reflecting increasing severity with time. Figure 3 shows the difference in the mean number of coded diseases between patients with a positive outcome and ones with a negative outcome. This difference between the two groups, which can be used as proxy for difference in measured disease severity, was highest among the tumor patients followed by COPD, CHF and dementia—which correlated directly with the AUC of the best-performing random forest model.

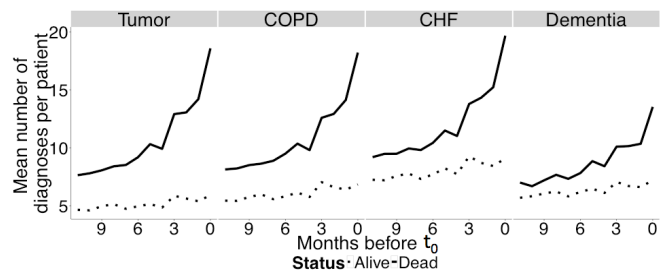

Fig. 2. Patterns in underlying disease severity & evolution

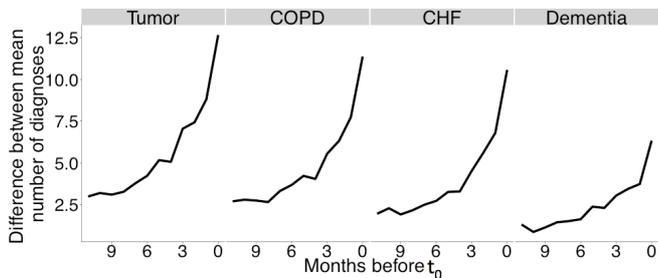

Fig. 3. Difference in disease severity between events and non-events

## V. Discussion

Using nationally-representative administrative claims of beneficiaries across a range of different disease types, we developed an augmented set of variables capturing different aspects of a patient's health condition. We screened multiple ML algorithms to find the best performing algorithm. We found that a RF classifier presented with an augmented set of variables and a balanced subsample to grow each tree i.e., a 'balanced RF' outperforms all other classifiers.

We observed large improvements in model performance with the augmented set of variables developed for this study, using a combination of domain knowledge and statistical cross-validation. This was likely the result of three key factors. First, these variables quantified disease severity by using counts of encounters rather than simply presence or absence of the medical encounter or disease. Second, they capture disease progression over time, by splitting the variables into two discrete periods. These two attributes lead to an improved ability to learn differences between patients with increases in healthcare utilization and disease severity near the end of life, and patients who show low utilization and disease severity levels. Third, we added commonly overlooked yet clinically significant variables, including patients' ability to function independently, which is known to be highly correlated to prognosis but ignored by traditional comorbidity measures.

Classifier performance was affected by the pattern of recorded comorbidity in the patients studied. While it is commonly believed that degree of class imbalance is a major determinant of classification performance [17], almost all the classifiers produced better predictions for the tumor cohort, despite the fact that it suffered from the highest class imbalance relative to all other disease cohorts. This was likely because the time trend and level of recorded diagnoses in patients who went on to die were quite different than patients who did not. In general, the bigger the difference in disease severity between the patients with the positive outcome and the ones with the negative outcome, the better the predictions were.

Our study had at least two major limitations. First, a dataset that accurately captures individual diagnoses and medical encounters of patients is a complex individual-level time series dataset, and creating a feature set from these fine-grained inputs requires a number of choices that are often somewhat arbitrary. There are about 25 million possible combinations[4] of diagnoses, dates, and encounter patterns and it would not be feasible to screen all possible features for inclusion in the model. Thus choices must be made, relying on content expertise rather than more formalized feature selection techniques. It is possible that a different set of variables, which incorporates different choices of variables, might have outperformed the proposed one. Future work could focus on formalizing these choices

Second, our results depend to a large extent on the ML screening process and the choice of tuning parameter values. It may be true that the screening algorithm we used did not find the globally optimal classifier, tuning parameters and class balancing techniques. However, since the RF family completely outperformed all others and variations within the same family of classifiers yield minimal changes in performance, we believe this model is the best classifier for this and similar large administrative claims datasets.

## VI. Conclusion

Accurate assessment of patients' likelihood to survive on the short run is crucial in guiding physician and patient decisions as well as population health management efforts. Prior studies have produced prognostic tools that are not practical in a real life setting because of technical and practical limitations. In this paper we used routinely collected administrative data to construct a unique feature set that captures disease severity and progression and screened the most widely used ML algorithms to create a prognostic tool that outperformed ones commonly used in medical literature. Eventually, similar classifiers could be used to feedback prognostic information to clinicians or policy-makers to identify patients at high risk of short-term mortality, and allow honest patient-physician discussions about end of life care and treatment options.

## References

[1] L. C. Yourman, S. J. Lee, M. A. Schonberg, E. W. Widera, and A. K. Smith, "Prognostic Indices for Older Adults: A Systematic Review," *JAMA: The Journal of the American Medical Association*, vol. 2, pp. 182-192, Jan. 2012.

[2] G. King, and L. Zeng, "Logistic Regression in Rare Events Data," *Political Analysis*, vol. 2, pp. 137-163, Feb. 2001.

[3] J. A. Cruz, and D. S. Wishart, "Applications of Machine Learning in Cancer Prediction and Prognosis". *Cancer Inform*, vol 2, pp. 59-77, Feb. 2011.

[4] G. Slaughter, Z. Kurtz, M. DesJardins, P. F. Hu, C. Mackenzie, L. Stansbury, and D.M. Stien, "Prediction of Mortality". *Proceedings of the ninth IEEE Biomedical Circuits and Systems Conference (BIOCAS)*, pp 1-4, Taiwan: IEEE press, 2012.

[5] S. McMillan, C. C. Chia, A. V. Esbroeck, I. Rubinfeld, and Z. Syed, "ICU Mortality Prediction using Time Series Motifs." *Proceedings of the 2012 Annual Computing in Cardiology Conference*. pp. 265-268. Poland, 2012.

[6] M. Ghassemi, T. Naumann, F. Doshi-Velez, N. Brimmer, R. Joshi, A. Rumshisky, and P. Szolovits, "Unfolding Physiological State: Mortality Modelling in Intensive Care Units." *Proceedings of the twentith ACM SIGKDD Conference on Knowledge Discovery and Data Mining*. pp 75-84. New York: ACM press, 2014.

[7] D. Bertsimas, M. V. Bjarnadottir, M. A. Kane, J. C Kryder, R. Pandey, S. Vempala, and G. Wang, "Algorithmic Prediction of Health-Care Costs". *Operations Research*, vol 6, pp 1382-1392, Nov. 2008.

[8] A. Hosseinzadeh, M. Izadi, A, Verma, D. Percup, and D. Buckeridge, "Assessing Predictability of Hospital Readmission Using Machine Learning". *Proceedings of the Twenty Fifth Innovative Applications of Artificial Intelligence Conference*. pp 1532-1538. Quebec City, Quebec: AAAI press, 2013

[9] P. Domingos, "A few useful things to know about machine learning." *Communications of the ACM*, vol. 55 (10), pp 78-87, Oct. 2012.

[10] J. J. Gagne, R. J. Glynn, J. Avorn, R. Levin, and S. Schneeweiss, 2011. "A combined comorbidity score predicted mortality in elderly patients better than existing scores." *Journal of Clinical Epidemiology*, vol 7, pp. 749-759, July 2011.

[11] J. E. Wennberg, D. O. Staiger, S. M. Sharp, D. J. Gottlieb, G. Bevan, K. Mcpherson, and H.G. Welch, "Observational intensity bias associated with illness adjustment: cross sectional analysis of insurance claims". *BMJ*, vol 346, Feb. 2013.

---

[4] Number is calculated by multiplying the number of possible ICD codes with the number of days in a year and the number of different medical encounters that patient can have.

[12] S. J. Lee, K. Lindquist, M.R. Segal, and K. Covinsky, "Development and Validation of a Prognostic Index for 4-Year Mortality in Older Adults". *JAMA: The Journal of the American Medical Association*, vol 7, pp. 801-808, Feb. 2006.

[13] S.K. Inouye, P. N. Peduzzi, J. T. Robison, J. S. Hughes, R. I. Horwitz, and J. Concato, "Importance of Functional Measures in Predicting Mortality Among Older Hospitalized Patients". *JAMA: The Journal of the American Medical Association*, vol. 15, pp. 1187-1193, April 1998.

[14] L. R. Shugarman, S. L. Decker, and A. Bercovitz, "Demographic and Social Characteristics and Spending at the End of Life". *Journal of Pain and Symptom Management*, vol. 38, pp 15-26, July 2009.

[15] P. Cunningham, M. Cord, and S. J. Delany, "Supervised Learning" in *Machine learning techniques for multimedia: case studies on organization and retrieval*. M. Cord and P. Cunningham, Eds. Berlin: Springer, 2008, ch. 2, pp. 21-49.

[16] S. B. Kotsiantis, "Supervised Machine Learning: A Review of Classification Techniques". *Informatica*, vol. 31, pp. 249-268, Nov. 2007.

[17] H. He, and E. A. Garcia, "Learning From Imbalanced Data" *IEEE Transactions on Knowledge and Data Engineering*, vol. 21, pp. 1263-1284, Sept. 2009